\documentclass[]{article}
\usepackage[letterpaper]{geometry}
\usepackage{mtsummit2015}

\usepackage{times}
\usepackage{url}
\usepackage{latexsym}
\usepackage{natbib}
\usepackage{layout}

\usepackage{amsmath}
\usepackage{multirow}
\usepackage{graphicx}
\usepackage{linguex}
\usepackage{amssymb}
\usepackage{amsmath}
\usepackage{colortbl}
\usepackage{arabtex}
\usepackage{amsfonts}
\usepackage{pifont}
\usepackage{algpseudocode}
\usepackage{algorithm}
\usepackage{rotating}
\usepackage{cjhebrew}

\newcommand{\hide}[1]{}

\newcommand{\mateana}{{\sc AMEANA}}

\newcommand{\gen}{{\sc Gen}}
\newcommand{\num}{{\sc Num}}
\newcommand{\dt}{{\sc Det}}

\newcommand{\pos}{{\sc Pos}}

\newcommand{{\scs}}{\sc SCS}
\newcommand{{\tcs}}{\sc TCS}

\newcommand{{\smw}}{\sc Ws}
\newcommand{{\tmw}}{\sc Wt}

{%

\hrulefill
\vspace*{0.5cm}%
\end{minipage}
}

{ \begin{list}%
	{$\bullet$}%
	{\setlength{\labelwidth}{25pt}%
	 \setlength{\leftmargin}{30pt}%
	 \setlength{\itemsep}{\parsep}}}%
{ \end{list} }

\renewcommand{\epsilon}{\varepsilon}

\begin{document}

%\mtsummitHeader{x}{x}{xxx-xxx}{2015}{45-character paper description goes here}{Author(s) initials and last name go here}
\title{Morphological Constraints \\ for Phrase Pivot Statistical Machine Translation}  

\author{\name{\bf Ahmed El Kholy} \hfill  \addr{ame2127@columbia.edu}\\ 
        \addr{Center for Computational Learning Systems, Columbia University \\}
\AND
       \name{\bf Nizar Habash} \hfill \addr{nizar.habash@nyu.edu}\\
        \addr{Computer Science, New York University Abu Dhabi}
}

\maketitle
\pagestyle{empty}

\begin{abstract}

 The lack of parallel data for many language pairs
  is an important challenge to statistical machine translation (SMT). One common
  solution is to pivot through a third language for which there exist
  parallel corpora with the source and target languages. Although
  pivoting is a robust technique, it introduces some low quality
  translations especially when a poor morphology language is used as the pivot between  rich morphology languages.  
  In this paper, we examine the use of synchronous morphology
  constraint features to improve the quality of phrase pivot SMT. 
  We compare hand-crafted constraints to those learned from limited parallel data between source and target languages.  
  The learned morphology constraints are based on projected alignments between the
  source and target phrases in the pivot phrase table. We show positive results on Hebrew-Arabic SMT (pivoting on English). We get 1.5 BLEU points over a phrase pivot  baseline and 0.8 BLEU points over a system combination baseline with a direct model built from  parallel data.
  \end{quote}
\end{abstract}

\parskip=0.00in

\setarab
\novocalize

\section{Introduction}
\label{intro} 

One of the main challenges in statistical machine translation (SMT) is the
scarcity of parallel data for many language pairs especially when the
source and target languages are morphologically rich.  
A common SMT solution to the lack of parallel data is to pivot the
translation through a third language (called pivot or bridge language)
for which there exist abundant parallel corpora with the source and
target languages. The literature covers many pivoting techniques. One
of the best performing techniques, phrase pivoting
\citep{utiyama-isahara:2007:main}, builds an induced new phrase table
between the source and target.  One of the main issues of this
technique is that the size of the newly created pivot phrase table is
very large. Moreover, many of the
produced phrase pairs are of low quality which affects the translation
choices during decoding and the overall translation quality.

In this paper, we focus on improving phrase pivoting. We introduce morphology constraint scores which are added to the log linear space of features in order to determine the quality of the pivot phrase pairs. We compare two methods of generating the morphology constraints. One method is based on hand-crafted rules relying on the authors knowledge of the source and target languages; while in the other method, the morphology constraints are induced from available parallel data between the source and target languages which we also use to build a direct translation model. We then combine both the pivot and direct models to achieve better coverage and overall translation quality. We show positive results on Hebrew-Arabic SMT. We get 1.5 BLEU points over a phrase-pivot baseline and 0.8 BLEU points over a system combination baseline with a direct model built from given parallel data.

Next, we briefly discuss some related work.  In
Section~\ref{pivot_strategy}, we  review the best performing
pivoting strategy and how we use it. In Section~\ref{ling}, we discuss the linguistic differences
among Hebrew, Arabic, and the pivot language, English. This is followed by our approach to
using morphology constraints in Section~\ref{approach}. We finally
present our experimental results in Section~\ref{experiments} and a case study in Section~\ref{case_study}.
  
\section{Related Work}
\label{related}

Many researchers have investigated the use of pivoting (or bridging)
approaches to solve the data scarcity issue
\citep{utiyama-isahara:2007:main,wu-wang:2009:ACLIJCNLP,Khalilov2008,bertoldi2008phrase,habash-hu:2009:WMT}. The
main idea is to introduce a pivot language, for which there exist
large source-pivot and pivot-target bilingual corpora.  
Pivoting has been explored for closely related languages
\citep{HHK:2000} as well as unrelated languages
\citep{koehn2009462,habash-hu:2009:WMT}.
Many different
pivot strategies have been presented in the literature. 
The following three are  the most common.
The first strategy is the sentence translation technique in which we
first translate the source sentence to the pivot language, and then
translate the pivot language sentence to the target language
\citep{Khalilov2008}.
The second strategy is based on phrase pivoting
\citep{utiyama-isahara:2007:main,cohn2007machine,wu-wang:2009:ACLIJCNLP}. In
phrase pivoting, a new source-target phrase table (translation model)
is induced from source-pivot and pivot-target phrase tables. Lexical
weights and translation probabilities are computed from the two
translation models. 
The third strategy is to create a synthetic source-target corpus by
translating the pivot side of source-pivot corpus to the target
language using an existing pivot-target model
\citep{bertoldi2008phrase}. 
In this paper, we use the phrase pivoting approach, which has
been shown to be the best with comparable settings
\citep{utiyama-isahara:2007:main}.

There has been recent efforts in improving phrase pivoting. One effort focused on improving alignment symmetrization targeting pivot phrase systems \citep{elalignment}. In another recent effort, Multi-Synchronous
Context-free Grammar (MSCFG) is leveraged to triangulate source-pivot
and pivot-target synchronous Context-free Grammar (SCFG) rule tables into a source-target-pivot
MSCFG rule table that helps in remembering the pivot during decoding. Also, pivot LMs are used to assess the naturalness of the derivation \citep{miura-EtAl:2015:ACL-IJCNLP}. 

In our own previous work, we demonstrated quality improvement using connectivity strength features between the source and target phrase pairs in the pivot phrase table \citep{elkholy-EtAl:2013:Short}. These features provide quality scores based on the number of alignment links between words in the source phrase to words of the target phrase. In this work, we extend on the connectivity scores with morphological constraints through which we provide quality scores based on the morphological compatibility between the connected/aligned source and target words.	

Since both Hebrew and Arabic are morphologically rich, we should
mention that there has been a lot of work on translation to and from
morphologically rich languages
\citep{yeniterzi:2010,elming-habash:2009:Semitic,Kholy:2010a,habash-sadat:2006:HLT-NAACL06-Short,kathol2008strategies}. Most
of these efforts are focused on syntactic and morphological processing
to improve the quality of translation.

Until recently, there has not been much parallel Hebrew-English and Hebrew-Arabic data
\citep{Tsvetkov:2010}, and consequently little work on Hebrew-English and Hebrew-Arabic
SMT.  \cite{Lavie:2004} built a transfer-based translation system
for Hebrew-English and so did \cite{shilon2012machine} for
translation between Hebrew and Arabic. Our previous work discussed above \citep{elkholy-EtAl:2013:Short} was demonstrated on Hebrew-Arabic with English pivoting.

\section{Phrase Pivoting}
\label{pivot_strategy}
\label{filtering}

\begin{table*}[!t]
%\begin{footnotesize}
\begin{center}
\begin{tabular}{|l|c|c|c|}
\hline
	& \textbf{Training Corpora} & \multicolumn{2}{c|}{\textbf{Phrase Table}}  \\\cline{3-4}	
\textbf{Translation Model}	&  \textbf{Size}	& \textbf{\# Phrase Pairs}	& \textbf{Size} \\ \hline
Hebrew-English	& 	$\approx$1M	words & 	3,002,887	&	327MB \\\hline
English-Arabic	& 	$\approx$60M	 words & 111,702,225	& 	14GB \\\hline
Pivot\_Hebrew-Arabic		& N/A	& 	$>$ 30 Billion 	& $\approx$2.5TB \\\hline
\end{tabular}
\end{center}
%\end{footnotesize}
\caption{Translation Models Phrase Table comparison in terms of number of lines and sizes.}
\label{sizes}
\end{table*}

In this section, we review the phrase pivoting strategy in detail as we describe how we built our baseline for Arabic-Hebrew via pivoting on English. We also discuss how we overcome the large expansion of source-to-target phrase pairs in the process of creating a pivot phrase table.  
In phrase pivoting (which is sometimes called triangulation or phrase table
multiplication), we train a Hebrew-Arabic and an English-Arabic
translation models, such as those used in the sentence pivoting
technique. Based on these two models, we induce a new Hebrew-Arabic
translation model.
Since our models are based on a Moses phrase-based SMT system \citep{Koehn07b}, we use the standard set of phrase-based translation probability
distributions.\footnote{Four different phrase translation scores are
  computed in Moses' phrase tables: two lexical weighting scores and two phrase
  translation probabilities.} We follow
\cite{utiyama-isahara:2007:main} in computing the pivot phrase pair
probabilities. The following are the set of equations used to compute the
lexical probabilities ($p_{w}$) and the phrase translation probabilities ($\phi$):
%%%%%%%%%%%%%%%%%%%%%%%%%%%%%%%%%%%%%%%%%%%%%%%%%%%%%%%%%%%
\begin{center}
$\phi(h|a) = \sum\limits_{e} \phi(h|e) \phi(e|a)$ \\
$\phi(a|h) = \sum\limits_{e} \phi(a|e) \phi(e|h)$ \\
$p_{w}(h|a) = \sum\limits_{e} p_{w}(h|e) p_{w}(e|a)$ \\
$p_{w}(a|h) = \sum\limits_{e} p_{w}(a|e) p_{w}(e|h)$
\end{center}
Above, $h$ is the Hebrew source phrase; $e$ is the English pivot phrase that is common in both Hebrew-English translation model and English-Arabic translation model; and $a$ is the Arabic target phrase.
%%%%%%%%%%%%%%%%%%%%%%%%%%%%%%%%%%%%%%%%%%%%%%%%%%%%%%%%%%%
%
We also build a Hebrew-Arabic reordering table using the same
technique but we compute the reordering probabilities in a similar manner to
\cite{marinolearning}.

\paragraph{Filtering for Phrase Pivoting}
As discussed earlier, the induced Hebrew-Arabic phrase and reordering
tables are very large. Table \ref{sizes} shows the amount of parallel
corpora used to train the Hebrew-English and the English-Arabic and
the equivalent phrase table sizes compared to the induced
Hebrew-Arabic phrase table.\footnote{The size of the induced phrase
  table size is computed but not created.} 
We follow the work of \cite{elkholy-EtAl:2013:Short} and filter the phrase pairs used in pivoting based on log-linear scores.
The main idea of the filtering process is to select the top [{\em n}]
English candidate phrases for each Hebrew phrase from the
Hebrew-English phrase table and similarly select the top [{\em n}]
Arabic target phrases for each English phrase from the English-Arabic
phrase table and then perform the pivoting process described earlier
to create a pivoted Hebrew-Arabic phrase table.
To select the top candidates, we first rank all the candidates based
on the log linear scores computed from the phrase translation
probabilities and lexical weights multiplied by the optimized decoding
weights then we pick the top [{\em n}] pairs. In our experiments, we pick the top 1000 pairs for pivoting.

\section{Linguistic Comparison}
\label{ling}

In this section we present the challenges of preprocessing
Arabic, Hebrew, and English, and how we address them.
Both Arabic and Hebrew are morphologically complex languages. 
One aspect of Arabic's complexity is its various attachable clitics
and numerous morphological features \citep{Habash-ANLP:2010}. 
Clitics include conjunction proclitics, e.g., +<w>~{\em w+}\footnote{Arabic transliteration throughout the paper is presented in the Habash-Soudi-Buckwalter scheme \citep{HSB-TRANS:2007}.}~`and',
  prepositional  proclitics, e.g., +<l>~{\em l+}~`to/for', the definite
  article +<Al> {\em Al+} `the', and the class of pronominal
  enclitics, e.g., <hm>+ {\em +hm} `their/them'. 
  All of these clitics are separate words in English.
  Beyond the 
clitics, Arabic words inflect for person, gender, number,
aspect, mood, voice, state and case.
Additionally, Arabic orthography uses optional diacritics for short vowels
and consonant doubling. This, together with Arabic's morphological richness, leads to 
a high degree of ambiguity: about 12 analyses per word,
typically corresponding to two lemmas on average
\citep{Habash-ANLP:2010}.
We follow \cite{Kholy:2010a} and use the PATB tokenization scheme
\citep{Maamouri:2004} in our experiments. 
The PATB scheme separates all
  clitics except for the determiner clitic {\em Al+}({\dt}).
We use MADA v3.1
\citep{Habash:2005,Habash:2009} to tokenize the Arabic text.  We only
evaluate on detokenized and orthographically correct (enriched) output
following the work of \cite{Elkholy:lrec2010}.

Similar to Arabic, Hebrew poses computational processing challenges typical of Semitic
languages \citep{hebrew-lr,shilon2012machine}.
Hebrew orthography also uses optional diacritics and its
morphology 
inflects for gender, number, person, state, tense and definiteness.
Furthermore, Similar to Arabic, Hebrew has a set of attachable clitics, e.g., conjunctions (such as +\<w>~{\it
  w+}\footnote{The following Hebrew 1-to-1 transliteration is
  used (in Hebrew alphabetical 
  order): \emph{abgdhwzxTiklmns`pcqr\v{s}t}. All examples are
  undiacritized and final forms are not distinguished from non-final
  forms.}~`and'), prepositions (such as +\<b>~{\it b+}~`in'), the definite
article (+\<h>~{\it h+}~`the'), or pronouns (such as \<hM>+~{\it
  +hm}~`their').  These issues contribute to a high degree of
ambiguity that is a challenge to translation from Hebrew to English or
to any other language. We follow \cite{singh2012hebrew}'s best preprocessing setup which  
 utilized a Hebrew tagger \citep{adler2007hebrew} and produced a tokenization scheme that separated all clitics.

English, our pivot language, is quite different from
both Arabic and Hebrew. English is poor in morphology and barely
inflects for number and tense, and for person in a limited
context. English preprocessing simply includes down-casing, separating
punctuation and splitting off ``'s''.

\section{Approach}
\label{approach}

One of the main challenges in phrase pivoting is the very large size
of the induced phrase table. It becomes even more challenging if
either the source or target language is morphologically rich. The
number of translation candidates (fanout) increases
%\hide{\footnote{fanout is the number of phrase pairs containing the
%    target phrase in a phrase-table})} 
due to ambiguity and richness
%(discussed in more details in Section~\ref{ling}) 
which in return
increases the number of combinations between source and target
phrases. Since the only criteria of matching between the source and
target phrase is through a pivot phrase, many of the induced phrase
pairs are of low quality. These phrase pairs unnecessarily increase
the search space and hurt the overall quality of translation. A basic solution to the combinatorial expansion
is to filter the phrase pairs used in pivoting based on log-linear scores as discussed in Section~\ref{filtering}, however, this doesn't solve the low quality problem. 

%To address the quality problem, 
Similar to factored translation models \citep{koehn2007factored} where linguistic (morphology) features are augmented to the translation model to improve the translation quality, our approach to address the quality problem is based on constructing a list of synchronous morphology constraints between the source and target languages. These constraints are used to generate scores to determine the quality of pivot phrase pairs. However, unlike factored models, we do not use the morphology in generation and the morphology information comes completely from external resources. In addition, since we work in the pivoting space, we only apply the morphology constraints to the connected words between the source and target languages through the pivot language. This guarantees a fundamental level of semantic equivalence before applying the morphology constraints especially if there is distortion between source and target phrases. 

We build on our approach in \cite{elkholy-EtAl:2013:Short} where we introduced connectivity strength features between the source and target phrase pairs in the pivot phrase table. These features provide quality scores based on the number of
alignment links between words in the source phrase and words in the target phrase. The alignment links are generated by projecting the alignments of the source-pivot phrase pairs and  the pivot-target phrase pairs used in pivoting. We use the same concept but instead of using the lexical mapping between source and target words, we compute quality scores based on the morphological compatibility between the connected source and target words. 

To choose which morphological features to work with, we
performed an automatic error analysis on the output of the phrase-pivot baseline
system. We did the analysis using  {\mateana} \citep{ameana}, an open-source  error analysis tool for natural language processing tasks targeting morphologically rich languages.
We found  that the most problematic morphological features in the Arabic output are gender ({\gen}), number ({\num}) and determiner ({\dt}). We focus on those features in addition to ({\pos}) in our experiments.

Next, we present our approach to generating the morphology constraint features using hand-crafted rules and compare this approach with  inducing these constraints from Hebrew-Arabic parallel data.

\subsection{Rule-based Morphology Constraints}
\label{rules}

%In this section, we explain our approach in generating morphology constraint features using hand-crafted rules. 
%These rules are basically a list of mappings of the different morphological features between Hebrew and Arabic. 
Our rule-based morphology constraint features are  
basically a list of hand-crafted mappings of the different morphological features between Hebrew and Arabic. 
Since both languages are morphologically rich as explained in  Section~\ref{ling}, 
it is straightforward to produce these mappings for {\gen}, {\num} and {\dt}. 
Note, however, that we also account for ambiguous cases; e.g., feminine gender in Arabic can map to words with ambiguous gender in Hebrew. We additionally use different {\pos} tag sets for Arabic (47 tags) and Hebrew (25 tags) and in many cases one Hebrew tag can map to more than one Arabic tag; for example, three Arabic noun tags {\it abbrev, noun} and {\it noun\_prop} map to two Hebrew tags {\it feminine, masculine} noun.\footnote{Please refer to \citep{Habash:2009} for a complete set of Arabic {\pos} tag set and \citep{adler2007hebrew} for Hebrew {\pos} tag set.} Table~\ref{rulesmap} shows a sample of the morphological mappings between Arabic and Hebrew.

\begin{table}[!ht]
\begin{center}
\begin{tabular}{|l|l|l|}
\hline
\bf Features & \bf Arabic & \bf Hebrew \\
\hline
{\gen}	&	Feminine	&	Feminine / Both 	\\ \cline{2-3} 
	&	Masculine	&	Masculine / Both	\\ \hline \hline
{\num}	&	Singluar &	Singluar / Singluar-Plural 	\\ \cline{2-3} 
	&	Dual 	&	Dual  / Dual-Plural	\\ \cline{2-3}  
	&	Plural 	&	Plural  / Dual-Plural / Singular-Plural	\\ \hline \hline
{\dt}	& No Determiner  &	No Determiner\\ \cline{2-3} 
		& Determiner		&	Determiner  \\ \hline
\end{tabular}
\end{center}
\caption{Rule-based mapping between Arabic and Hebrew morphological features. Each feature value in Arabic can map to more than one feature value in Hebrew.}
\label{rulesmap}
\end{table}

After building the morphological features mappings, we use them to judge the quality of a given phrase pair in the phrase pivot model.  We add two scores $W_{s}$ and $W_{t}$ to the log linear space. 
%Given a {\it source,target} phrase pair $\bar{s},\bar{t}$ and a word projected alignment $ a $ between the source word positions $i=1, ..., n$ and the target word positions $j=1,...,m$, $W_{s}$ and $W_{t}$ are defined in equations~\ref{eq:SCS}
%and \ref{eq:TCS} where $ F $ is the set of morphological features (we focus on {\gen}, {\num}, {\dt} and {\pos}) and 
%$M_{f}$ is the hand-crafted rules mapping feature values of feature $f \in F$  and $MLE_{f}(i)$ is the maximum likelihood feature value of feature $f$ for the source word at position $i$ while $MLE_{f}(j)$ is the maximum likelihood feature value of feature $f$ for the target word at position $j$.
Given a {\it source-target} phrase pair $\bar{s},\bar{t}$ and a word projected alignment $ a $ between the source word positions $i=1, ..., n$ and the target word positions $j=1,...,m$, $W_{s}$ and $W_{t}$ are defined in equations~\ref{eq:SCS} and \ref{eq:TCS}.
$ F $ is the set of morphological features (we focus on {\gen}, {\num}, {\dt} and {\pos}). 
$M_{f}$ is the hand-crafted rules mapping between Arabic and Hebrew feature values of feature $f \in F$. In case of ambiguity for a given feature; for example, a word's gender being masculine or feminine, we use the maximum likelihood value of this feature given the word.
$MLE_{f}(i)$ is the maximum likelihood feature value of feature $f$ for the source word at position $i$, and  $MLE_{f}(j)$ is the maximum likelihood feature value of feature $f$ for the target word at position $j$. The maximum likelihood feature values for Hebrew were computed from the Hebrew side of the training data. As for Arabic, the  maximum likelihood feature values  were computed from the Arabic side of the training data in addition to Arabic Gigaword corpus, which was used in creating the language model (more details in Section~\ref{setup}).

\begin{center}
\begin{small}
\begin{equation} \label{eq:SCS}
W_{s}  = \frac{1}{|F|} \sum\limits_{\forall f \in F} \sum\limits_{\forall (i,j) \in a }\frac{1}{n}[(MLE_{f}(i),MLE_{f}(j)) \in M_{f}]  
\end{equation}
\end{small}
\end{center}

\begin{center}
\begin{small}
\begin{equation} \label{eq:TCS}
W_{t}  = \frac{1}{|F|} \sum\limits_{\forall f \in F} \sum\limits_{\forall (i,j) \in a }\frac{1}{m}[(MLE_{f}(i),MLE_{f}(j)) \in M_{f}]  
\end{equation}
\end{small}
\end{center}

\subsection{Induced Morphology Constraints}
\label{morph}

In this section, we explain our approach in generating morphology constraint features from a given parallel data between source and target languages. Unlike the rule-based approach we build a translation model between the source and target morphological features and we use the morphology translation probabilities as metric to judge a given phrase pair in the pivot phrase table. For the automatically induced constraints, we jointly model mapping between conjunctions of features attached to aligned words rather than tallying each feature match independently. Writing good manual rules for such feature conjunction mappings would be more difficult. Table~\ref{induced_example} shows some examples of mapping ({\gen}), number ({\num}) and determiner ({\dt}) in Hebrew to their equivalent in Arabic and their respective bi-directional scores.

\begin{table}[ht]
\begin{center}
\begin{tabular}{|l|l|c|c|}
\hline
\bf Hebrew (H) & \bf Arabic (A) & \bf $P_{FC}(A|H)$ & \bf $P_{FC}(H|A)$ \\
\hline
%[Fem+Dual+Det]	&	[Fem+Dual]	&	0.00056243	& 0.0833333 \\ \hline
%[Fem+Dual+Det]	&	[Fem+Dual+Det]	&	0.0148148	&	0.333333 \\ \hline
%[Fem+Dual+Det]	&	[Fem+Singular] [Fem+Dual]	&	0.00518135	&	0.0833333 \\ \hline
%[Fem+Dual+Det]	&	[Masc+Dual+Det]	&	0.00466563	&	0.5 \\ \hline
[Fem+Dual+Det]	&	[Fem+Dual]	&	0.0006	&	0.0833	\\ \hline
[Fem+Dual+Det]	&	[Fem+Dual+Det]	&	0.0148	&	0.3333	\\ \hline
[Fem+Dual+Det]	&	[Fem+Singular] [Fem+Dual] &	0.0052	&	0.0833	\\ \hline
[Fem+Dual+Det]	&	[Masc+Dual+Det] &	0.0047	&	0.5000	\\ \hline
\end{tabular}
\end{center}
\caption{Examples of induced morphology constraints for ({\gen}), number ({\num}) and determiner ({\dt}) and their respective scores.}
\label{induced_example}
\end{table}

As in rule-based approach, we add two scores $W_{s}$ and $W_{t}$ to the log linear space which are defined in equations~\ref{eq:SCSM} and \ref{eq:TCSM}. $P_{FC}$ is the conditional morphology probability of a given feature combination $(FC)$ value. Similar to rule-based morphology constraints, we resort to the maximum likelihood value of a feature combination when the values are ambiguous.
 %given a certain word which is inherently less common. 
 $MLE_{FC}(i)$ is the maximum likelihood feature combination $(FC)$ value for the source word at position $i$ while $MLE_{FC}(j)$ is the maximum likelihood feature combination $(FC)$ value for the target word at position $j$.

\begin{center}
\begin{small}

\begin{equation} \label{eq:SCSM}
W_{s}  = \frac{1}{n} \sum\limits_{\forall (i,j) \in a} P_{FC}(MLE_{FC}(i) | MLE_{FC}(j))  
\end{equation}
\end{small}
\end{center}

\begin{center}
\begin{small}

\begin{equation} \label{eq:TCSM}
W_{t}  = \frac{1}{m} \sum\limits_{\forall (i,j) \in a} P_{FC}(MLE_{FC}(j) | MLE_{FC}(i))  
\end{equation}
\end{small}
\end{center}

\subsection{Model Combinations}
\label{modelcomb}

%Since we use  parallel data to induce the morphology constraint rules, it would make sense to see the effect of combining the pivot model with the added morphology constraints and the direct model trained of the parallel data used to induce the rules. We use Moses phrase table combination techniques \citep{koehn2007experiments} and we show results over a learning curve as shown in Section~\ref{curve_section}.
%

Since we use parallel data to induce the morphology constraints, it would make sense to measure the effect of combining (a) the pivot model with added morphology constraints, and (b) the direct model trained on the parallel data used to induce the morphology constraints. We perform the combination using Moses' phrase table combination techniques. Translation options are collected from one table, and additional options are collected from the other tables. If the same translation option (in terms of identical input phrase and output phrase) is found in multiple tables, separate translation options are created for each occurrence, but with different scores \citep{koehn2007experiments}. We show results over a learning curve in Section~\ref{curve_section}.

\section{Experiments}
\label{experiments}

In this section, we present a set of experiments comparing the use of rule-based versus induced morphology constraint features in  phrase-pivot SMT as well as model combination to improve Hebrew-Arabic pivot translation quality.

\subsection{Experimental Setup}
\label{setup}

In our pivoting experiments, we build two SMT models; one model to
translate from Hebrew to English, and another model to translate from
English to Arabic. The English-Arabic parallel corpus is about ($\approx$ 60M words) and is available from LDC\footnote{LDC Catalog
  IDs: LDC2005E83, LDC2006E24, LDC2006E34, LDC2006E85, LDC2006E92,
  LDC2006G05, LDC2007E06, LDC2007E101, LDC2007E103, LDC2007E46,
  LDC2007E86, LDC2008E40, LDC2008E56, LDC2008G05, LDC2009E16,
  LDC2009G01.} 
  and GALE\footnote{Global Autonomous Language
  Exploitation, or GALE, was a DARPA-funded research project.}
constrained data. 
The Hebrew-English corpus is  about ($\approx$ 1M words) and is available from sentence-aligned corpus produced by \cite{Tsvetkov:2010}.
For the direct Hebrew-Arabic SMT model, we use a TED parallel corpus of about ($\approx$ 2M words) \citep{cettoloEtAl:EAMT2012}.
%\footnote{TED data can be download from: https://wit3.fbk.eu/mt.php?release=2012-02}.

%
Word alignment is done using GIZA++ \citep{och:03syst}.
%We compare two alignment
%techniques, one based on the surface form of the Arabic word and the
%other based on the Arabic lemma. Results are discussed in the next section.
%
For Arabic language modeling, we use 200M words from the Arabic
Gigaword Corpus \citep{Graff:2007} together with the Arabic side of our
training data. We use 5-grams for all language models (LMs) implemented using the SRILM
toolkit \citep{Stolcke:2002}. 

%For English language modeling, we use
%English Gigaword Corpus with 5-gram LM using the KenLM toolkit
%\citep{heafield2011kenlm}.

%The same preprocessing was used on the English data for all experiments. 

%
 
All experiments are conducted using the Moses phrase-based SMT system
\citep{Koehn07b}. We use MERT \citep{och2003minimum} for decoding weight
optimization. Weights are
optimized using a tuning set of 517 sentences developed by \cite{shilon2010machine}.

We use a maximum phrase length of size 8 across all models.  We report
results on a Hebrew-Arabic development set (Dev) of 500 sentence with a single reference and an evaluation set (Test) of 300 sentences with three references developed by \cite{shilon2010machine}. We evaluate using BLEU-4 \citep{BLEU}.

%\subsection{Pivoting}

\subsection{Baselines}
\label{baseline_eval}

We compare the performance of adding the connectivity
strength features ({\em +Conn}) to the phrase pivoting SMT model ({\em Phrase\_Pivot}) and building a direct SMT model using all parallel He-Ar corpus available.
The results are presented in Table~\ref{baselines}. Consistently with our previous effort \citep{elkholy-EtAl:2013:Short}, the performance of the phrase-pivot model improves with the connectivity strength features. While the direct system is better than the phrase pivot model in general, the combination of both models leads to a high performance gain of 1.7/4.4 BLEU points in Dev/Test over the best performers of both the direct and phrase-pivot models.

\begin{table}[ht]
\begin{center}
\begin{tabular}{|l|c|c|}
\hline
\bf Model & \bf Dev & \bf Test \\
\hline
Direct & 9.7 & 20.4 \\ \hline \hline
Phrase\_Pivot & 8.3 & 19.8  \\ \hline 
Phrase\_Pivot+Conn & 9.1 & 20.1  \\ \hline \hline
Direct+Phrase\_Pivot+Conn & \textbf{11.4} & \textbf{24.5} \\ \hline
\end{tabular}
\end{center}
\caption{Comparing phrase pivoting SMT with connectivity strength features, direct SMT and the model combination. The results show that the best performer is the model combination in Dev and Test sets.}
\label{baselines}
\end{table}

\subsection{Rule-based Morphology Constraints}

\begin{table*}[ht]
\begin{center}
\begin{tabular}{|l|c|c|c|c|}
\hline
\bf Model & \multicolumn{2}{c|}{\textbf{Dev}} & \multicolumn{2}{c|}{\textbf{Test}} \\ \cline{2-5}
 & \bf Single & \bf Combined & \bf Single  & \bf Combined \\ \hline
Direct & \textbf{9.7} &  n/a & 20.4 &  n/a \\ \hline \hline
Phrase\_Pivot+Conn & 9.1 & 11.4 & 20.1 & 24.5 \\ \hline
Phrase\_Pivot+Conn+Morph\_Rules & \textbf{9.6} & 12.2\** & 20.9\** &  24.6 \\ \hline
Phrase\_Pivot+Conn+Morph\_Auto & \textbf{9.6} & \textbf{12.4}\** & \textbf{21.6}\** & \textbf{25.3}\** \\ \hline

\end{tabular}
\end{center}
\caption{Morphology constraints results. The ``Single" columns show the results of a single model of either the direct model or the phrase pivoting models with additional morphological constraints features. The ``Combined" show the results of system combination between the direct model and the different phrase pivoting models. In the first row, the ``Combined" results are not applicable for the direct model. (*) marks a statistically significant result against both the direct and phrase-pivot baseline.}
\label{morph_results}
\end{table*}

In this experiment, we show the performance of adding hand-crafted morphology constraints ({\em +Morph\_Rules}) to determine the quality of a given phrase pair in the phrase-pivot translation model. The third row in Table~\ref{morph_results} shows that although the rules are based on a one-to-one mapping between the different morphological features, the translation quality is improved over the baseline phrase-pivot model by 0.5/0.8 BLEU points in Dev/Test sets.

%Despite the fact that the performance hasn't improved compared to the direct model in the Dev set, the Test set showed a great improvement of 0.8 and 1.2 BLEU points over the phrase-pivot and direct models respectively. This is could be due to the effect of having single reference compared to multi-reference evaluation sets. 

As expected, the system combination of the pivot model with the direct model improves the overall performance but  the gain we get from the morphology constraints only appears in the Dev set with 0.8 BLEU points, and not much in the Test set.

%\begin{table}[!ht]
%\begin{center}
%\begin{tabular}{|l|c|c|c|c|}
%\hline
%\bf Model & \bf Dev & \bf Test \\ \hline
%Direct & 9.7 & 20.4 \\ \hline
%Phrase\_Pivot+Conn & 9.1 & 20.1  \\ \hline
%Phrase\_Pivot+Conn+Morph\_Rules & \textbf{9.6} &  20.9 \\ \hline
%\end{tabular}
%\end{center}
%\caption{Rule-based morphology constraints.}
%\label{rules_results}
%\end{table}

\subsection{Induced Morphology Constraints}

In this experiment, we measure the effect of using induced morphology constraints ({\em +Morph\_Auto}) 
on MT quality.
%to determine the quality of a given phrase pair in the pivot translation model. 
The last row in Table~\ref{morph_results} shows that the induced morphology constraints improve the results over the baseline phrase-pivot model by 0.5/1.5 BLEU points in Dev/Test sets and over the Rule-based morphology constraints by 0.7 BLEU points in the Test set.

Similar to the Rule-based constraints, the performance did not improve compared to the {\it direct model} in the Dev set; but, again, the Test set showed a great improvement of 1.5 and 1.2 BLEU points over the pivot and direct models, respectively. Also the system combination of the pivot model with the direct model improves the overall performance. The  model using induced morphological features is the best performer with an increase in the performance gain by 1.0/0.8 BLEU points in Dev/Test sets. This shows that the benefit we get from the induced morphology constraints were not diluted when we do the model combination given the fact that the constraints were induced from the parallel data to start with.

It is important to note here that the induced morphology constraints outperformed the rule-based constraints across all settings. This shows that the complex morphology constraints extracted from the parallel data provide knowledge that can not be covered by simple linguistic rules. However, the simple rule-based approach comes in handy when there is no data between the source and target languages.

%\begin{table}[!ht]
%\begin{center}
%\begin{tabular}{|l|c|c|}
%\hline
%\bf Model & \multicolumn{2}{c|}{\textbf{Test}} \\ \cline{2-3}
%		  & \bf Single & \bf Comb \\ \hline
%Direct & \textbf{9.7} &  n/a \\ \hline
%Phrase\_Pivot+Conn+Morph\_Rules &  20.9 &  24.6 \\ \hline
%Phrase\_Pivot+Conn+Morph\_Auto & \textbf{21.6} & \textbf{25.3} \\ \hline
%\end{tabular}
%\end{center}
%\caption{Rule-based morphology constraints.}
%\label{rules_comb_results}
%\end{table}

\subsection{Learning Curve}
\label{curve_section}

In this experiment, we examine the effect of using less data in inducing morphology constraints rules and the overall performance when we combine systems. Table \ref{curve_results} shows the results on a learning curve of 100\% (2M words), 25\% (500K words) and 6.25\% (125K words) of the parallel Hebrew-Arabic corpus. 

\begin{table*}[!t]
\setlength{\tabcolsep}{4pt}
\begin{center}
\begin{tabular}{|c|l|c|c|c|c|}
\hline
\bf Parallel & \bf Model & \multicolumn{2}{c|}{\textbf{Dev}} & \multicolumn{2}{c|}{\textbf{Test}} \\ \cline{3-6}
\bf Data Size &  & \bf Single & \bf Combined& \bf Single & \bf Combined \\ \hline
\textbf{125K} & Direct & 2.7 &  n/a & 8.4 &  n/a \\ \cline{2-6}
 			& Phrase\_Pivot+Conn & 9.1 & 10.4 & 20.1 & 20.9\\ \cline{2-6}
			& Phrase\_Pivot+Conn+Morph\_Auto & 9.2 & 10.6 & 20.6 &	21.3 \\ \hline \hline
\textbf{500K} & Direct & 5.9 &  n/a & 15.1 &  n/a \\ \cline{2-6}
 			& Phrase\_Pivot+Conn & 9.1	& 10.7 & 20.1 & 22.5\\ \cline{2-6}
			& Phrase\_Pivot+Conn+Morph\_Auto & 9.7	& 11.2 & 20.8 &	22.8 \\ \hline \hline
\textbf{2M} & Direct & \textbf{9.7} &  n/a & 20.4 &  n/a \\ \cline{2-6}
 			& Phrase\_Pivot+Conn & 9.1 & 11.4 & 20.1 & 24.5\\ \cline{2-6}
			& Phrase\_Pivot+Conn+Morph\_Auto & \textbf{9.6} & \textbf{12.4} & \textbf{21.6} & \textbf{25.3} \\ \hline
\end{tabular}
\end{center}
\caption{Learning curve results of 100\% (2M words), 25\% (500K words) and 6.25\% (125K words) of the parallel Hebrew-Arabic corpus.}
\label{curve_results}
\end{table*}

As expected, The system combination between the direct translation models and the phrase-pivot translation model leads to an improvement in the translation quality across the learning curve even when there is small amount of parallel corpora. Despite the weak performance (2.7 BLEU) of the direct system built on 6.25\% of the parallel Hebrew-Arabic corpus, the system combination leads to 1.4 BLEU points gain.

An interesting observation from the results is that we always get a performance gain from the induced morphology constrains across all settings. This shows that the system combination helps in adding more lexical translation choices while the constraints help in a different dimension, which is selecting the best phrase pairs from the pivot system.

\section{Case Study}
\label{case_study}

In this section we consider an example from our Dev set that  captures many of the patterns and themes in the
evaluation.
% Test set -- see 
Table~\ref{casestudy} shows a Hebrew source sentence  and its Arabic reference. This is followed by the output from the pivot system, the direct system, the Phrase\_Pivot+Conn+Morph\_Auto system and the combined system. 

Two particular aspects should be noted. First is the complementary lexical coverage of the direct and pivot systems. This is seen in how one of each covers half of the phrase {\it middlemen and traders}. The combined system captures both.  Secondly, the gender, number and tense of the main verb prove challenging in many ways (and this is an issue for a majority of the sentences in the Dev set).  The Hebrew verb in the present tense is masculine and plural; and naturally follows the subject.  The Arabic reference verb  appears at the beginning of the sentence, in which location it only agrees with the subject in gender (while number is singular).  
Arabic Verbs in SVO order agree in gender and number. 
All the MT systems we compare leave the verb after the subject. The direct, Phrase\_Pivot+Conn+Morph\_Auto, and combination  systems get the  number and gender correctly; however, the direct and combined system make the verb tense past.    The Phrase\_Pivot+Conn+Morph\_Auto example highlights the value of morphology constraints; but the example points out that they sometimes are hard to evaluate automatically, since there are morphosyntactically allowable forms that do not match the translation references.

\begin{table*}
\setlength{\tabcolsep}{2pt}
\begin{center}
\begin{footnotesize}
\begin{tabular}{|l|l|}
\hline

\textbf{Hebrew Source} & \multicolumn{1}{r|}{.\<hmtwwkym whswxrym msrbym ldbr bpwmby  `l hmxyrym>} \\ 
& {\it the+middlemen and+the+traders refuse[m.p.] to+speak publicly about the+prices}\\\hline
\textbf{Arabic Reference} &  \multicolumn{1}{r|}{<yrf.d Alws.tA' wAlt^gAr Al.hdy_t `lnA `n AlAs`Ar>} \\
& {\it refuse[m.s.] the+middlemen and+the+traders the+speaking publicly about the+prices}\\\hline\hline
\textbf{Phrase\_Pivot+Conn} &  \multicolumn{1}{r|}{ <yrf.d Alt.hd_t `lnA `n AlAs`Ar> \<whswxrym> <ws.tA'> } \\
& {\it middlemen \<whswxrym> refuse[m.s.]  the+speaking publicly about the+prices}\\\hline

\textbf{Direct} &  \multicolumn{1}{r|}{ <wAlt^gAr rf.dwA Al.hdy_t `lY Almla' `lY  AlAs`Ar> \<hmtwwkym>} \\
& {\it \<hmtwwkym> and+the+traders refused[m.p.]  the+speaking upon the+public about the+prices}\\\hline

\textbf{Phrase\_Pivot+Conn+}&  \multicolumn{1}{r|}{ <yrf.dwn Alt.hd_t `lnA `n AlAs`Ar> \<whswxrym> <alws.tA'> } \\
\verb|    |\textbf{Morph\_Auto}& {\it the+middlemen \<whswxrym> refuse[m.p.]  the+speaking publicly about the+prices}\\\hline

\textbf{Direct+Phrase\_Pivot+}  &  \multicolumn{1}{r|}{<ws.tA' wAlt^gAr rf.dwA Alt.hd_t `lnA `n AlAs`Ar>} \\
\verb|    |\textbf{Conn+Morph\_Auto}& {\it middlemen and+the+traders refused[m.p.]  the+speaking publicly about the+prices}\\\hline

\end{tabular}
\end{footnotesize}
\end{center}
\caption{Translation examples. }
\label{casestudy}
\end{table*}

\section{Conclusion and Future Work}
\label{conclusions}

In this paper, we presented the use of synchronous morphology
  constraint  features based on hand-crafted rules compared to rules induced from parallel data 
  to improve the quality of phrase-pivot based SMT. We show that the two approaches lead to an improvement in the translation quality. The induced morphology constraints approach is a better performer, however, it relies on the fact there is a parallel corpus between source and target languages. We show positive results on Hebrew-Arabic SMT. We get 1.5 BLEU points over phrase-pivot baseline and 0.8 BLEU points over system combination baseline with direct model built from given parallel data.
  
   In the future, we plan to work on reranking experiments as a post-translation step based on morphosyntactic information between source and target languages. We also plan to work on word reordering between morphologically rich language to maintain the relationship between the word order and the morphosyntactic agreement in the context of phrase pivoting.

\section*{Acknowledgments}
The work presented in this paper was possible thanks to a generous 
Google Research Award.   We would like to thank Reshef Shilon and Shuly Winter  
for helpful discussions and support with processing Hebrew.
We also thank the anonymous reviewers for their insightful comments.

%\section*{Acknowledgments}
%The work presented in this paper was possible thanks to a generous research grant from Google
%We also thank the anonymous reviewers for their insightful comments.

\small

\bibliography{MTSummit2015.bbl}

%\bibliographystyle{acl}
%\bibliography{ALLBIB-1.8,newbib}

\end{document}